\documentclass[conference]{IEEEtran}
\IEEEoverridecommandlockouts
\usepackage{cite}
\usepackage{tablefootnote}
\usepackage{amsmath,amssymb,amsfonts}
\usepackage{algorithmic}
\usepackage{graphicx}
\usepackage{textcomp}
\usepackage{xcolor}
\def\BibTeX{{\rm B\kern-.05em{\sc i\kern-.025em b}\kern-.08em
    T\kern-.1667em\lower.7ex\hbox{E}\kern-.125emX}}
\begin{document}

\title{Real-time Addressee Estimation: Deployment of a Deep-Learning Model on the iCub Robot \\
\thanks{For research leading to these results Carlo Mazzola has received funding from the project titled TERAIS in the framework of the program Horizon-Widera-2021 of the European Union under the Grant agreement number 101079338.}
}

\author{\IEEEauthorblockN{1\textsuperscript{st} Carlo Mazzola}
\IEEEauthorblockA{\textit{CONTACT unit} \\
\textit{Istituto Italiano di Tecnologia}\\
Genoa, Italy \\
carlo.mazzola@iit.it}
\and
\IEEEauthorblockN{2\textsuperscript{nd} Francesco Rea}
\IEEEauthorblockA{\textit{CONTACT unit} \\
\textit{Istituto Italiano di Tecnologia}\\
Genoa, Italy \\
francesco.rea@iit.it}
\and
\IEEEauthorblockN{3\textsuperscript{rd} Alessandra Sciutti}
\IEEEauthorblockA{\textit{CONTACT unit} \\
\textit{Istituto Italiano di Tecnologia}\\
Genoa, Italy \\
alessandra.sciutti@iit.it}
}

\maketitle

\begin{abstract}
Addressee Estimation is the ability to understand to whom a person is talking, a skill essential for social robots to interact smoothly with humans. In this sense, it is one of the problems that must be tackled to develop effective conversational agents in multi-party and unstructured scenarios. As humans, one of the channels that mainly lead us to such estimation is the non-verbal behavior of speakers: first of all, their gaze and body pose. Inspired by human perceptual skills, in the present work, a deep-learning model for Addressee Estimation relying on these two non-verbal features is designed, trained, and deployed on an iCub robot. The study presents the procedure of such implementation and the performance of the model deployed in real-time human-robot interaction compared to previous tests on the dataset used for the training.
\end{abstract}

\begin{IEEEkeywords}
Social Robots, Human-Robot Interaction, Deep Learning, Conversational Agents, Human Activity Recognition
\end{IEEEkeywords}

\section{Introduction} \label{Intro}
For artificial agents to be effective and smooth during conversational scenarios, multiple problems must be tackled. Focusing on the perceptual domain, i.e., a passive agent listening to humans, the artificial agents must be able to detect voices (Sound Detection and Voice Recognition), recognize who is talking (Speaker Recognition and Speaker Localization), and what they are saying (Natural Language Understanding). But even considering optimal performances in all these tasks, an artificial agent endowed with such abilities would hardly be able to cope with real-life environments. User Experiences with Conversational Agents are typically designed as one-to-one interactions. Even when groups are considered, interactions are conceived as agent-centered, as if artificial agents were always the intended interlocutors of speakers. But quite often, this is not the case of what happens outside laboratories.

Addressee Estimation (AE) is the ability to understand to whom the speaker is talking \cite{Skantze2020}. It represents an additional task, crucial for robots to interact with humans in unstructured environments. Thanks to this ability, robots can be enabled to 1) know if and when others address them, 2) understand social dynamics of communication (e.g., relations, roles, engagement of people, inclusion and exclusion processes, etc.), 3) contextualize the message communicated by the speaker based on its addressee. 

Humans can understand the addressee of an utterance via multiple channels. Verbal information provides some cues and context, but it has been demonstrated that non-verbal (and para-verbal) signals are crucial to achieving a correct estimation \cite{Skantze2020, Auer2018gaze, Ishii2016}. Adopting a human-inspired approach, visual information about the speakers' gaze, pose, and motion should be considered key inputs to developing AE models for robots and conversational agents. Previous works often focused on datasets without considering the implementation of models on artificial agents \cite{Jovanovic2006, Frampton2009, opdenakker2009, Malik2019, Malik2021, Le2018}. The ones deploying their model on an interactive agent mainly designed binary models to allow artificial agents only to estimate whether or not they were being addressed \cite{Bakx2003, Turnhout2005, Katzenmaier2004, Huang2011, Sheikhi2013} or relied on structured scenarios \cite{johansson2015opportunities}. 

Aiming at implementing AE skills in robots to let them interact in unstructured scenarios, this paper 1) describes the development of an AE deep-learning model trained on human-robot interaction (HRI) dataset, as already described in \cite{Mazzola2023}, 2) illustrates its first deployment on the humanoid robot iCub, and 3) reports the results of an HRI pilot experiment to evaluate the performance of the model deployed on the iCub compared to previous tests made on the training dataset.  

\section{Methods} \label{Methods}

This study tackles AE in multi-party HRI. The approach followed consists of developing and deploying a deep neural network (DNN) to solve the problem as addressee localization, more specifically, as a classification task based on the speakers' face and body pose. The design and training of the DNN are briefly described in Section \ref{Design}, summarizing \cite{Mazzola2023}. Thereafter, the architecture for deploying the model on the iCub robot and the pilot test of the deployed model are illustrated in Section \ref{Deployment}.  

\subsection{Design and training of the AE model} \label{Design}

Gaze and body pose represent two fundamental sources for AE. For this reason, we relied on them to design a model extracting such information from visual input, processing them in temporal sequences, and eventually providing a 3-class classification in terms of addressee localization. From the first-person perspective of the robot looking at the speaker, our model is designed to output the position of the addressee in three ways: the addressee is either at the robot's left, at the right, or the addressee is the robot.  

A DNN was developed to solve this task. It consists of a convolutional part to extract visual information from the speaker's face and body pose separately. Then, after combining the two modalities together, LSTM cells are meant to process temporal information. A last LogSoftMax layer provides the final classifaction and the confidence of the estimation (see \cite{Mazzola2023}). 

Collecting visual inputs with a frame rate of 12.5 Hz, the model is fed with input sequences of 10 frames, each sequence lasting 0.8 s, to balance fast with reliable estimations. In this way, the model provides a first estimate less than 1 second after the speaker starts to talk, granting rapid response for real-time interactions. Moreover, since utterances may be composed of several sequences, estimates of sequences belonging to the same utterance are balanced to provide a final estimate of the utterance's addressee.  

To gain ecological validity to the model, we chose the Vernissage Corpus \cite{Vernissage2013} for the training. Jayagopy et al. collected this dataset using a multi-party HRI scenario, with visual information recorded from a Nao robot's cameras. Conversations were designed so that each participant talked to the robot or another person in the room. The dataset creators had manually labeled the addressee of each utterance. As a plus point, the scenario was designed so that the speaker often talked to the addressee about objects in the environment, a typical feature of social communication called 'triadic interaction.' In this way, as it happens outside a lab, people not only look at each other during the conversation, but their gaze and body pose are affected by the target object \cite{Skantze2020, Auer2018gaze, Ishii2016}. For more information about the training procedure, see \cite{Mazzola2023}. 

\subsection{Deploying the AE model on the iCub robot} \label{Deployment}

The above-mentioned procedure has been adopted for the model to be deployed on a real robot. Once the model had been trained on the whole Corpus, it was ported on an iCub robot to endow it with real-time AE skills. 

The robot's architecture designed for the model to work in real-time consisted of 1) a visual sensory module to take in input information from one of the robot's cameras, 2) a controller for the robot's gaze, steering the iCub's neck and eyes 3) a face detector and an object tracker module to let the robot track who speak in case they move around, 4) a feature extractor, to compute the speakers' body pose (using a lightweight model \cite{osokin2018lightweight_openpose} optimizing OpenPose \cite{Openpose2019}), crop their face from the whole image, and create the sequences needed as input of the AE model. As a final step, the AE model classifies the addressee's location from the robot's first-person perspective \footnote{the code is available at the following link \\ https://gitlab.iit.it/cognitiveInteraction/addressee\_estimation\_irim2023.git}.

Although the architecture developed so far fully complies with the pipeline of the model and works in real-time, presently, it does not grant an autonomous AE skill to the robot involved in multi-party conversational scenarios. It partially requires the intervention of the programmer to select the speaker and trigger the model at the start of the utterance. Such a Wizard-of-Oz method may offer some advantages for the testing phase of the AE model, as is the case of the present study, or for collecting new data to improve the model performance. In any case, these two abilities, voice detection and voice localization, have already been developed on our iCub robot \cite{Gonzalez2021localization}. As a future step, they will be included in the architecture to achieve a fully autonomous HRI social skill. 

To test the model performance in real-time HRI, a pilot experiment was carried out with 6 volunteer participants (3 females and 3 males) interacting with the robot in pairs.  Participants gave their written informed consent before participating, and the regional ethical committee approved the study (Comitato Etico Regione Liguria). The position of participants was set to match the one adopted in the Vernissage Corpus \cite{Vernissage2013}. Participants were asked to talk freely to the other human or the robot and to comment on some objects in the environment at least twice to test the model on triadic interactions.

\section{Results} \label{Results}

\begin{figure}[b!]
\centerline{\includegraphics[trim={0 0.5cm 0.5cm 2cm},clip, width=1\columnwidth]{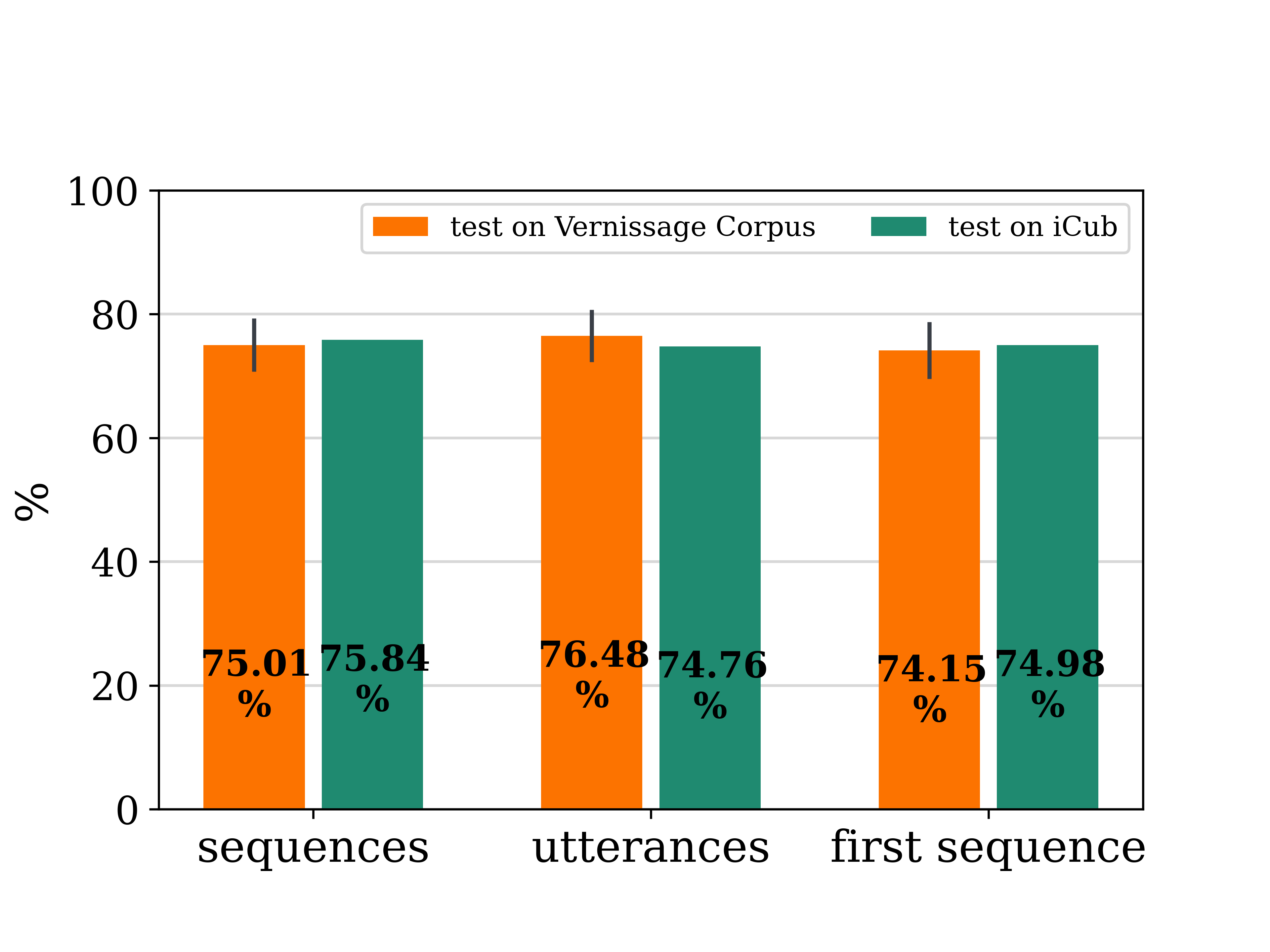}}
\caption{Bar plot comparing the performance of the AE model tested on the Vernissage Corpus with the one deployed on the iCub robot. The orange bars on the left represent the average F1-score from testing the Vernissage Corpus model with a 10-fold cross-validation technique (error bar represent SD). The green bars on the right represent the F1-score measured by testing the model in real-time on the iCub robot.}
\label{fig:barplot}
\end{figure}

Tests of the AE model on the Vernissage Corpus \cite{Vernissage2013} have been conducted using a 10-fold cross-validation technique. The model performance in terms of F1-score is, on average, 75.01\%, if computed on sequences, 76.48\% when aggregating sequences belonging to the same utterance, and 74.15\% if considering only the estimate of the first sequence of each utterance.

\begin{figure}[t!]
\centerline{\includegraphics[trim={0 0.2cm 4.5cm 0.5cm},clip, width=0.8\columnwidth]{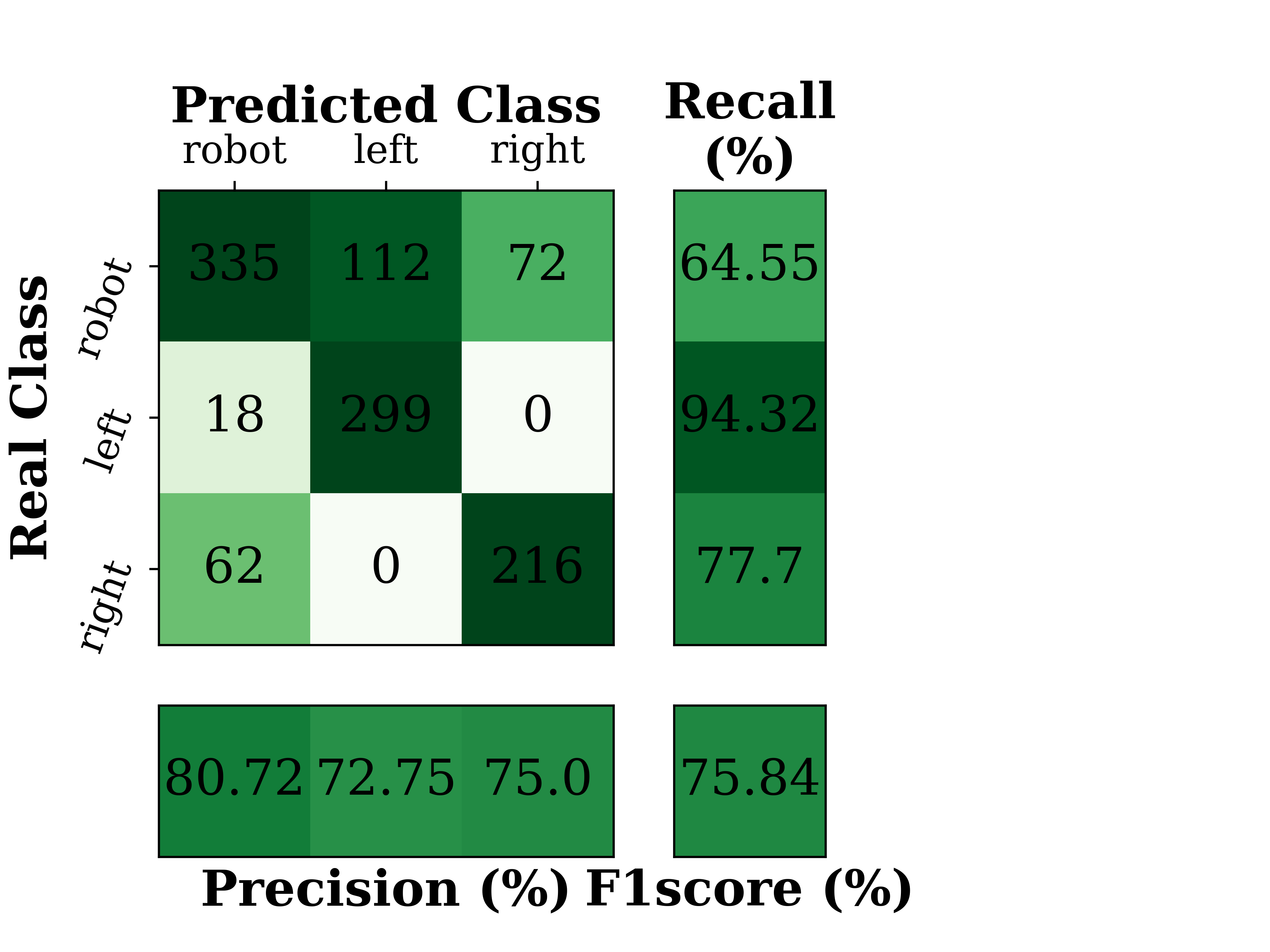
}}
\caption{Confusion Matrix resulting from the real-time test of the AE model deployed on the iCub. The matrix reports the performance of the model tested on sequences.}
\label{fig:matrix}
\end{figure}

This work aims to compare the model performance on the dataset with the performance of the model deployed on the iCub for real-time HRI. Figure  \ref{fig:barplot} compares the AE model performance on the Venissage Corpus and the iCub. In this last case, considering the same tests conducted on the dataset, the AE model exhibits an F1-score of 75.84\% on sequences, 74.76\% on sequences belonging to the same utterance, and 74.98\% at the first sequence of each utterance.

\begin{figure}[b!]
\centerline{\includegraphics[trim={2.5cm 1.5cm 2cm 1cm},clip, width=1\columnwidth]{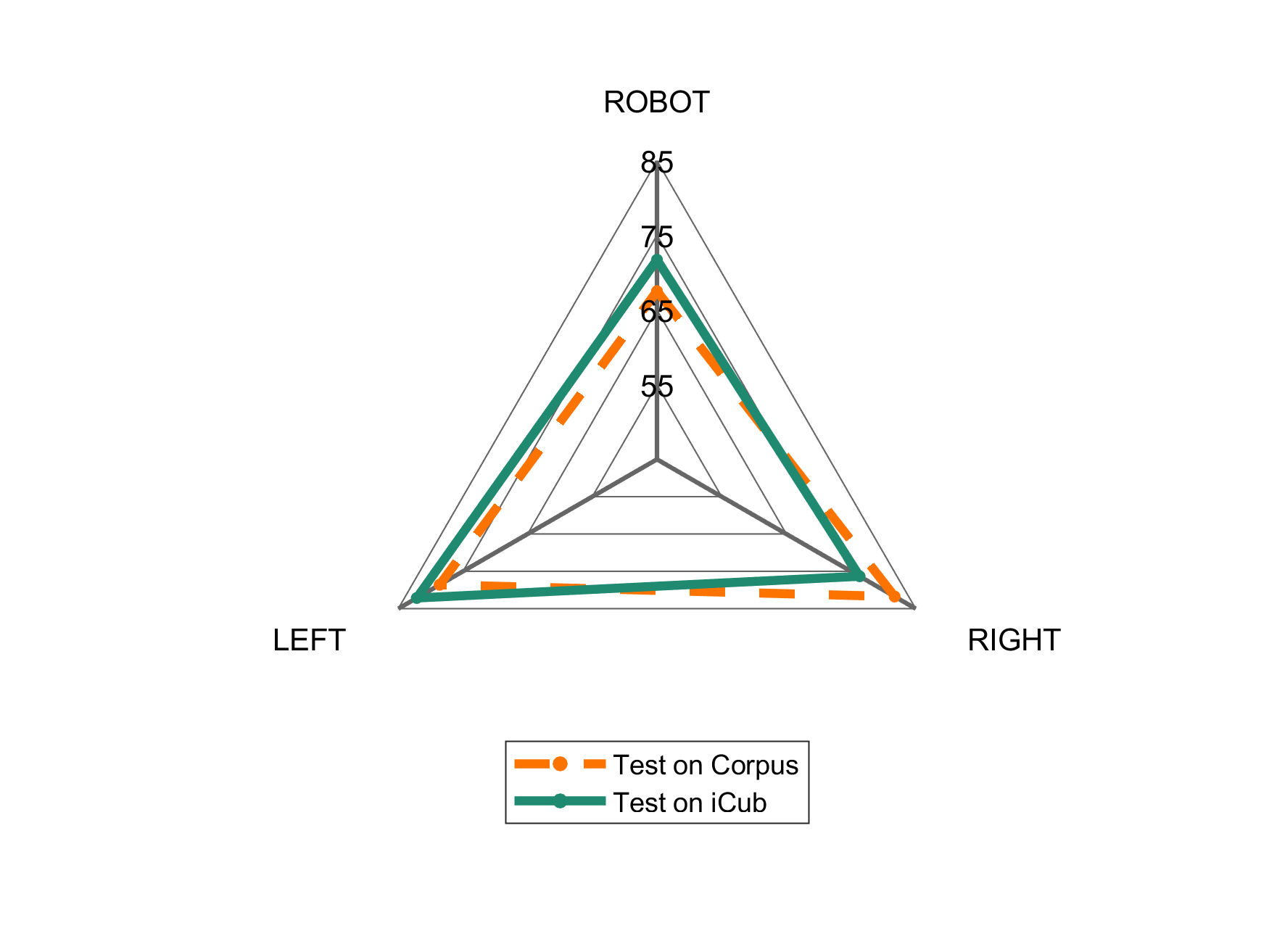}}
\caption{Spider plot comparing the performance of the AE model tested on the Vernissage Corpus with the one deployed on the iCub robot with a focus on each class. The plot reports the performance of the model tested on sequences.}
\label{fig:spider}
\end{figure}

The confusion matrix in Figure \ref{fig:matrix} displays more in detail the performance of the model deployed on the iCub for each class: 'robot', 'left', and 'right'. Moreover, with the same focus on classes, Figure \ref{fig:spider} provides a comparison with the previous test on the dataset. Whereas the weighted F1-score computed on the dataset amounts respectively to 67.52\%, 78.61\%, and 81.75\% for the 'robot', 'left', and 'right' classes. The F1-score of the model deployed on the iCub was 71.73\%, 82.14\%, and 76.33\% for the same classes.

\section{Discussion} \label{Discussion}

Addressee Estimation is the ability to understand to whom a person is talking. This work aims to develop this skill on the humanoid robot iCub to make it interact smoothly and in real-time in multi-party scenarios. To achieve this, a deep-learning model has been designed and trained \cite{Mazzola2023} on the Vernissage Corpus \cite{Vernissage2013} and implemented afterward on an iCub robot.

Results from a pilot experiment replicating some features of the Vernissage Corpus scenario revealed that the performance in real-time HRI with iCub is in line with the one assessed on the dataset (see Fig. \ref{fig:barplot}). Such results may be partially explained by the fact that the two scenarios were somehow similar, with people standing in front of the robot at a distance ranging between 1.5 to 2.5 meters. Nevertheless, one could expect a lower performance for the model deployed on the iCub because of several differences between the two conditions, such as the robot employed (Nao -- iCub), the experimental room, and the brightness. From this pilot experiment, it seems, therefore, that the model developed by \cite{Mazzola2023} has a decent degree of generalizability, allowing its exploitation in real-time interaction. 

The focus on classes provides additional proof of the model's performance (see Fig. \ref{fig:spider}). From this pilot test, it appears that on the iCub, the model gains a higher score when the speaker addresses the robot but loses some points when the addressee is at the right of the robot. However, more tests are needed to establish this pattern. 

Although the deployment of the model delivers a performance more satisfactory than expected, the analysis of the confusion matrix in Fig. \ref{fig:matrix} suggests an improvement, in particular regarding the estimation of the 'robot' class. To achieve this, future works will concern the collection of data, specific to the iCub, to enhance the performance with transfer learning techniques and make its deployment more effective.

Addressee Estimation is a core ability to make robots able to communicate smoothly with humans in multi-party and unstructured scenarios. Its implementation is required for conversational agents, which need non-verbal skills such as AE to support verbal abilities like Natural Language Understanding. But more generally, AE is essential for every robot interacting with humans. Features like the reception of even simple verbal or non-verbal commands, the correct interpretation of a human-populated environment, and the social interaction with every kind of human require robots to be aware of other agents' communicative intentions.

\section*{Acknowledgment}

We thank Marta Romeo and Angelo Cangelosi for their precious help in the design and training of the model and Pia Saade for her support in some steps of the deployment.

\bibliographystyle{IEEEtran}
\bibliography{IRIM-Addressee-2023}

\end{document}